\documentclass[10pt, a4paper]{article}
\usepackage{lrec}
\usepackage{multibib}
\newcites{languageresource}{Language Resources}
\usepackage{graphicx}
\usepackage{tabularx}
\usepackage{soul}
\usepackage{epstopdf}
\usepackage[utf8]{inputenc}

\usepackage{hyperref}
\usepackage{xstring}

\usepackage{color}

\title{Ethically Collecting Multi-Modal Spontaneous Conversations with People that have Cognitive Impairments}

\name{Angus Addlesee, Pierre Albert}

\address{Heriot-Watt University, University of Edinburgh\\
         Edinburgh, UK, Edinburgh UK\\
         ja204@hw.ac.uk, pierre.albert@ed.ac.uk\\}

\abstract{
In order to make spoken dialogue systems (such as Amazon Alexa or Google Assistant) more accessible and naturally interactive for people with cognitive impairments, appropriate data must be obtainable. Recordings of multi-modal spontaneous conversations with vulnerable user groups are scarce however and this valuable data is challenging to collect. Researchers that call for this data are commonly inexperienced in ethical and legal issues around working with vulnerable participants. Additionally, standard recording equipment is insecure and should not be used to capture sensitive data. We spent a year consulting experts on how to ethically capture and share recordings of multi-modal spontaneous conversations with vulnerable user groups. In this paper we provide guidance, collated from these experts, on how to ethically collect such data and we present a new system - ``CUSCO'' - to capture, transport and exchange sensitive data securely. This framework is intended to be easily followed and implemented to encourage further publications of similar corpora. Using this guide and secure recording system, researchers can review and refine their ethical measures. \\ \newline \Keywords{ethical data collection, multi-modal interaction, data security} 
}

\begin{document}

\maketitleabstract

\section{Introduction} \label{sec:intro}

In this paper, we first introduce the background and motivations behind our work before detailing our contributions to ethical protocols in section \ref{sec:ethics} We provide guidance, collated from meetings with experts, on ethically collecting multi-modal spontaneous conversations with people that have cognitive impairments. We have also created a system to securely record this data. This new system is detailed in section \ref{sub:device}

\subsection{Dialogue as Cognition Declines} \label{sub:dialogue}

Natural face-to-face conversations involve quick exchanges that are littered with hesitations, restarts, self-corrections \cite{Shriberg96disfluencies,Hough15}, interruptions \cite{Healey.etal11}, backchannels \cite{Heldner.etal13,howes2017feedback} and split utterances \cite{Howes12}, etc... with none of these phenomena respecting the boundaries of a sentence or turn. These phenomena become even more common and more pronounced as cognition declines. For example, people with certain types of dementia pause more frequently and for longer durations than healthy controls \cite{boschi2017connected}. These changes have even been used successfully to detect Alzheimer's disease (AD) from just a person's speech \cite{luz2018method,zhu2018detecting}.

People don't just communicate using words however, visual feedback (nodding, brow furrowing, head tilting, etc...) and hesitation utterances (``umm'', ``err'', ``hmm'', etc...) are short but do guide conversation \cite{charles1981conversational,bavelas2011listener}. Whether non-verbal interactions change as cognition declines is relatively unknown because multi-modal recordings of such interactions are scarce.

\subsection{Older Adults \& Spoken Dialogue Systems}

When people speak to Spoken Dialogue Systems (SDSs), such as Amazon Alexa or Google Assistant, they adapt to the system \cite{pelikan2016nao,Porcheron.etal18}. Each utterance is stripped of the phenomena discussed in section \ref{sub:dialogue} Adapting to an SDS is acceptable for the majority of users but older adults, with less exposure to such systems, and people with cognitive impairments can struggle to adapt their natural interaction patterns.

The global ageing population \cite{un:pop} relies on care systems that have been strained for years \cite{carersuk}. This causes knock-on problems, such as bed-blocking in hospitals \cite{beds}, and these pressures can be eased if people are able to live in their own homes for longer and more independently. A huge range of IoT devices could help tackle this challenge but their embedded SDSs need to become more natural if they are to make an impact \cite{sakakibara2017generating,helal2019smart}.

\subsection{Our Corpus Collection Details}

\begin{figure}[htb]
	\centering
	\includegraphics[width=1\linewidth]{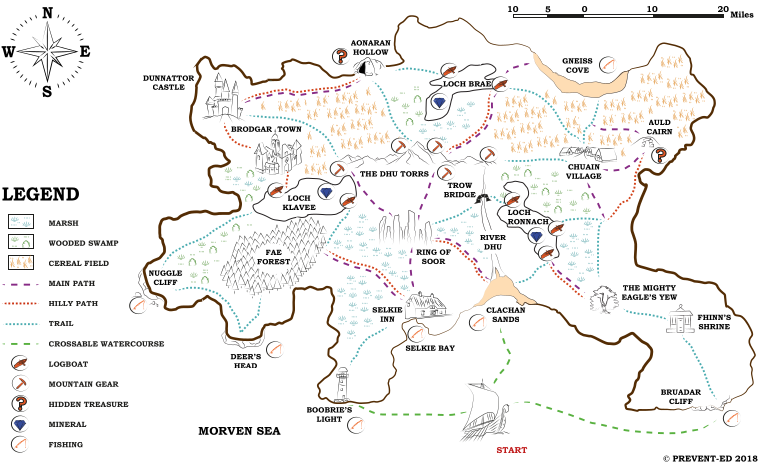}
	\caption{PREVENT-Elicitation of Dialogues (PREVENT-ED) map with routes \protect\cite{de2019protocol}.}
	\label{fig:map}
\end{figure}

For context, we are collecting a corpus of conversations with people that have various types of dementia. This does not constrain our work however and the ethical protocol extends to cover cognitive impairments more generally, established in Section \ref{sec:ethics}

Dementia is one of the leading causes of death in the UK \cite{ARUK2018deaths} but there is no treatment to prevent, cure or slow its progression. Even with suitable data, adapting SDSs for those living with dementia is not trivial due to the many challenges left to tackle \cite{addlesee2019current}. Monologue recordings of people with AD describing pictures exist \cite{becker1994natural} and interviews of people with AD exist \cite{pope2011finding} which are both used to develop the AD detection models mentioned in section \ref{sub:dialogue} Neither of these corpora contain the spontaneous conversational speech that we would expect an SDS to receive and, importantly, they are only audio recordings. Therefore, we are going to collect a multi-modal corpus of spontaneous conversations with people that have various types of dementia.

A variant of the map task \cite{anderson1991hcrc} has been recently developed to elicit spontaneous spatial navigation dialogues with people that have dementia \cite{de2019protocol} and we are collaborating with the creators of this task to collect our corpus. Using this task, a healthy participant will sit opposite a person that has dementia for a casual conversation. Both participants have a map with the same locations, but only the person with dementia can see the possible routes through the imaginary land (as shown in Figure \ref{fig:map}). The healthy participant, however, is the only one who knows which locations the pair need to visit. They therefore need to collaborate through conversation to go on the journey together.

\section{Ethical Considerations} \label{sec:ethics}

Over the past year, we have contacted and met with many experts to ensure that we collect our corpus ethically. These experts belong to many institutions including: The NHS, Alzheimer Scotland, Edinburgh Medical School, Edinburgh Centre for Dementia Prevention, Heriot-Watt University and more. We have collated all the information in this paper and we have also developed a new system to record multi-modal interactions more securely. This system is detailed in section \ref{sub:device}

\subsection{Consent} \label{sub:consent}

\begin{figure}[htb]
	\centering
	\includegraphics[width=1\linewidth]{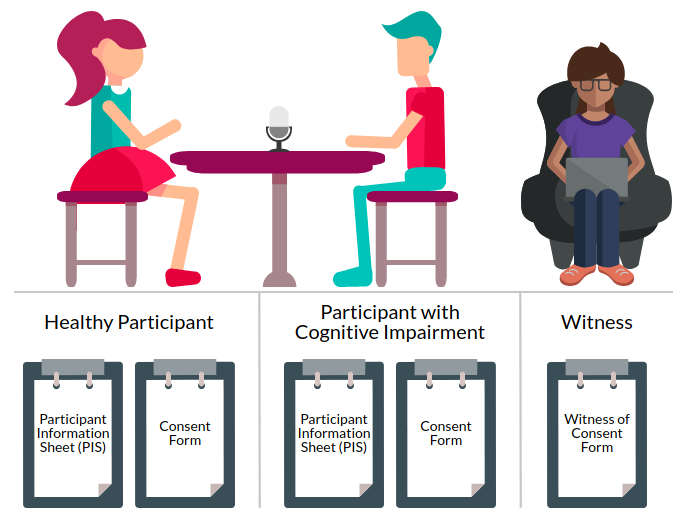}
	\caption{The distribution of required documentation.}
	\label{fig:docs}
\end{figure}

Each participant will be given a participant information sheet (PIS) before taking part in the study. This PIS contains all information about the study (what it will involve, the benefits of taking part, what data will be stored, etc...) and should be given to the participant at least a week before they take part. This time allows the participant to digest the information and ask any questions to family members, carers, GPs, or a member of the research team. A consent form is then provided before the experiment that summarises the key points in the PIS and confirms that the participant has read and understood it. It is important to stress that all questions are welcome and that participation is entirely voluntary. The documents are distributed as shown in Figure \ref{fig:docs}.

People with cognitive impairments are considered vulnerable participants and a witness is therefore required. The witness should be a family member or carer \cite{lacny2012predictors} and has to sign a witness of consent form. This should be signed \textit{after} the participant signs their consent form as it confirms that they understood the PIS, had all of their questions answered, and willingly consented to take part in the study. Immoral researchers could attempt to trick a person with a cognitive impairment (for example, offering to make them a cup of tea after they ``sign a quick form'') or elicit personal information (for example, asking about their previous medical history). To ensure this cannot happen, the witness also signs to confirm that the researcher did not attempt to elicit personal information, mislead, or trick the participant.

\subsection{Participant Comfort} \label{sub:comfort}

Participants are spending their valuable time helping with research but could feel stressed about taking part, especially those with cognitive impairments. Ensuring people have a comfortable experience is therefore of paramount importance.

Even before taking part, the PIS should contain as much information as possible to prevent unnecessary stress. For example, it can highlight the following about the task:

\begin{itemize}
    \setlength\itemsep{0em}
    \item It requires no preparation.
    \item It is not a medical examination.
    \item We want to record a natural conversation, so it is intended to be a fun game.
    \item Recording can be stopped (or paused) at any time without giving a reason.
    \item There is no right or wrong answer.
    \item There is no time limit.
\end{itemize}

Some people may feel uncomfortable stopping the study, even if they are feeling distressed. A family member or carer should witness the task for this reason, usually the same witness that we discussed in section \ref{sub:consent} The witness can also stop or pause the recording at any point without giving a reason. As a researcher, it is crucial to understand the importance of this witness. Different cognitive impairments and even different people with the same cognitive impairment have distinct signals to indicate distress. Family members and carers are significantly more experienced at identifying whether a particular person is uncomfortable, than any researcher, because they know that exact person.

\begin{figure}[htb]
	\centering
	\includegraphics[width=1\linewidth]{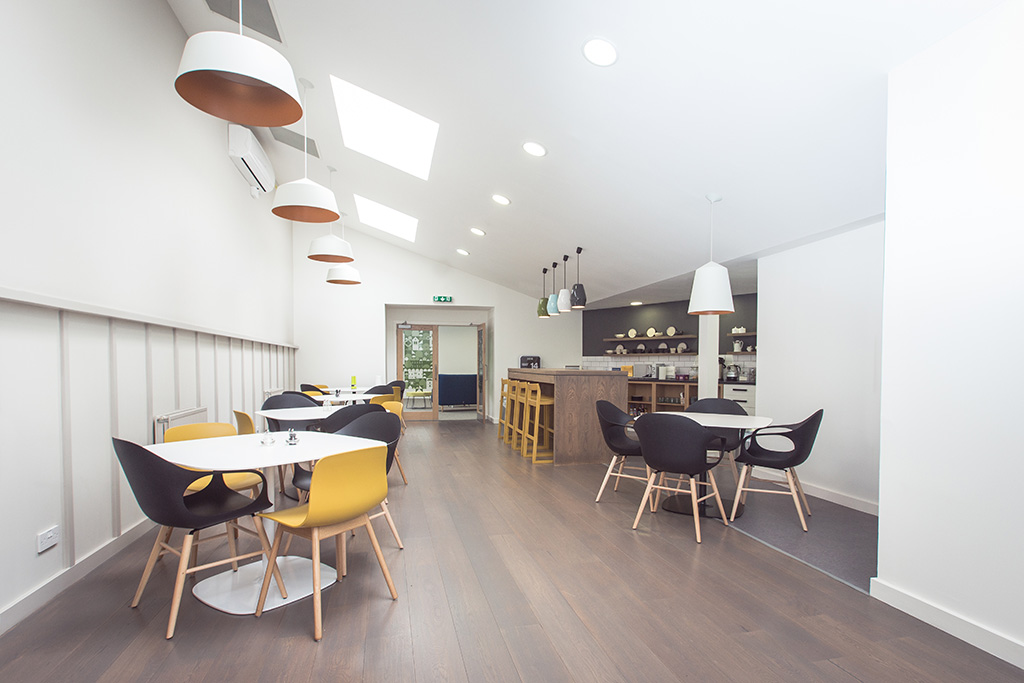}
	\caption{An Alzheimer Scotland café, designed to be an accessible community hub \protect\cite{graven2014centre}.}
	\label{fig:centre}
\end{figure}

A suitable location is relatively easy to find as spontaneous conversations can take place almost anywhere. For participant comfort and availability of a witness however, it is best to collaborate with a business or charity focused on the cognitive impairment of interest. To engage with their communities, these organisations usually have drop-in centres that people can visit for social activities, support, and classes (an example is shown in Figure \ref{fig:centre}). These centres are perfect locations to run tasks as people are very comfortable in them. The staff also know the potential participants and can therefore be witnesses and help with recruitment, discussed in section \ref{sub:recruit} Most accessible locations are suitable but working with a charity, to carry out the study in one of their centres, is the best option when possible.

\subsection{Participant Recruitment} \label{sub:recruit}

Collaborating with a relevant organisation is vital when recruiting vulnerable participants, this is in addition to the benefits around participant comfort. These organisations can reach out to their community and assist with recruitment of suitable participants in a safe and friendly manner. Collaboration costs the organisation valuable time however, so it is important to explain the motivations behind the data collection. We have had very positive responses from multiple charities using the rationale given in section \ref{sec:intro}

Healthy participants are also required to partake as interlocutors. It is common to compensate research participants with small rewards, such as gift cards, but it is not advised in this case. People with cognitive impairments will be using their time to contribute to research by taking part in the task. The healthy participant will ideally be motivated by the contribution to society and not some end reward. Someone who does not care about the motivations behind the research could rush through the task for a gift card, devaluing the vulnerable participants time. This example case can not happen if there is no monetary reward offered for taking part.

\subsection{Optional Cognitive Assessment}

Collecting multi-modal recordings of these conversations is a long and costly process. It is therefore worthwhile to share this data; and we will do so as detailed in section \ref{sub:data}. For use by other researchers in certain fields, such as Psychology, cognitive assessment results have huge benefits. For example, another corpus that performed the same cognitive assessment could be merged to reveal unknown connections. There are also worries to consider before including such a task however.

\begin{figure}[htb]
	\centering
	\includegraphics[width=1\linewidth]{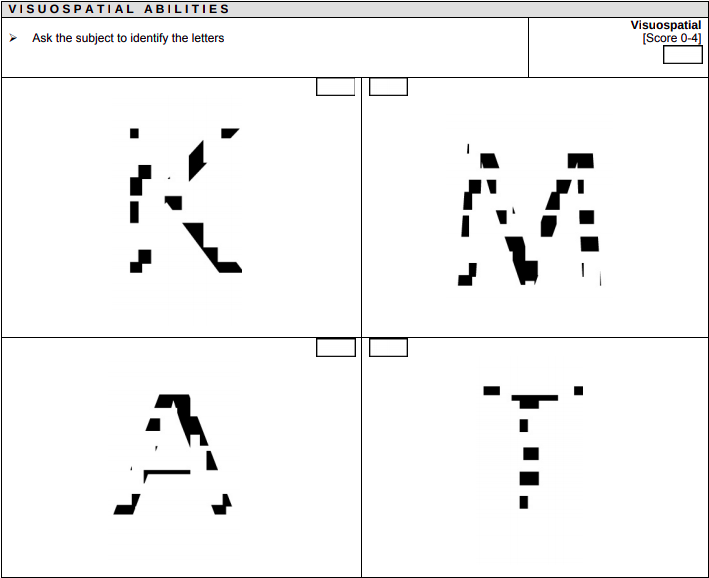}
	\caption{An example question from the ACE-III.}
	\label{fig:ace}
\end{figure}

For example, the task that is most suitable for our data collection is the Addenbrooke's Cognitive Examination (ACE-III) \cite{hsieh2013validation} as it is commonly used, low-tech, and quick to perform. An example question can be seen in Figure \ref{fig:ace}. Importantly, NHS training needs to be passed in order to run this test and it should not be recorded audiovisually. Similarly for other tests, all training must be completed prior to collection. One downside to highlight is that the ACE-III is used by GPs to screen people while diagnosing dementia. Therefore, participants may recall doing this task and be reminded of the stressful times around their diagnosis. This could upset a participant and in addition, retaking the test may highlight how they have declined in cognitive performance since first completing it.

Each cognitive impairment has a range of tests to scrutinise and it is very valuable to include a cognitive assessment. A person's well-being should be prioritised however, so only run tests after careful consideration and the relevant training.

\subsection{Securely Recording Multi-Modal Interactions} \label{sub:device}

Conversations involving patients or medical personnel are full of sensitive data. People often disclose personally identifiable information during a conversation (for example, mentioning their children's names or medical history). This concern is even stronger with conversations involving vulnerable participants (e.g. people with cognitive impairments), less prone to control the information they disclose. Standard recording systems (e.g. audio recorders and video cameras) are not secure devices, and they cannot be used to capture sensitive data. Furthermore, recorded data can easily be accessed on standard systems. Ethical and legal consequences of data breach must be accounted for if a standard device is lost or stolen, highlighting the need for a secure approach.

A new system - ``CUSCO'' - was developed to satisfy the requirements stemming from the ethical assessment regarding data collection of sensitive material. The device allows the collection of a range of modalities, including audio and video.
Handling conversations containing sensitive material requires mitigation of the consequences of unintentional or fraudulent loss of data. The device ensures the security of the recorded data by encrypting recorded streams in real time. The encryption is done using Veracrypt, a dedicated open-source software that underwent a security audit, vouching for the correct implementation of the encryption algorithms.

Collected data can only be accessed with the key generated for each project, ensuring security of the corpus during \textit{all} the phases of its life: collection, transport, exchange and storage.

The CUSCO device was designed to collect medical conversations between healthcare professionals and patients \textit{in-situ}. Recording material in medical practices is common to study real-life phenomena \cite{Montague_Asan_2014}, but - to our knowledge - considerations for the security of the collected data are overwhelmingly ignored.

Recent legal evolution on the protection of personal data, such as the General Data Protection Regulation (GDPR - \cite{gdpr}) in the EU, has led to a strong focus being put on these considerations during ethical evaluation and validation of new research projects involving data collection, use, and sharing.

Risk prevention and mitigation for data handling set the functional requirements for the design of our system. As such, even if the device is compromised or stolen \textit{during} recording, the entire dataset of previously recorded conversations and any recording in progress are secure.

\begin{figure}[htb]
	\centering
	\includegraphics[width=1\linewidth]{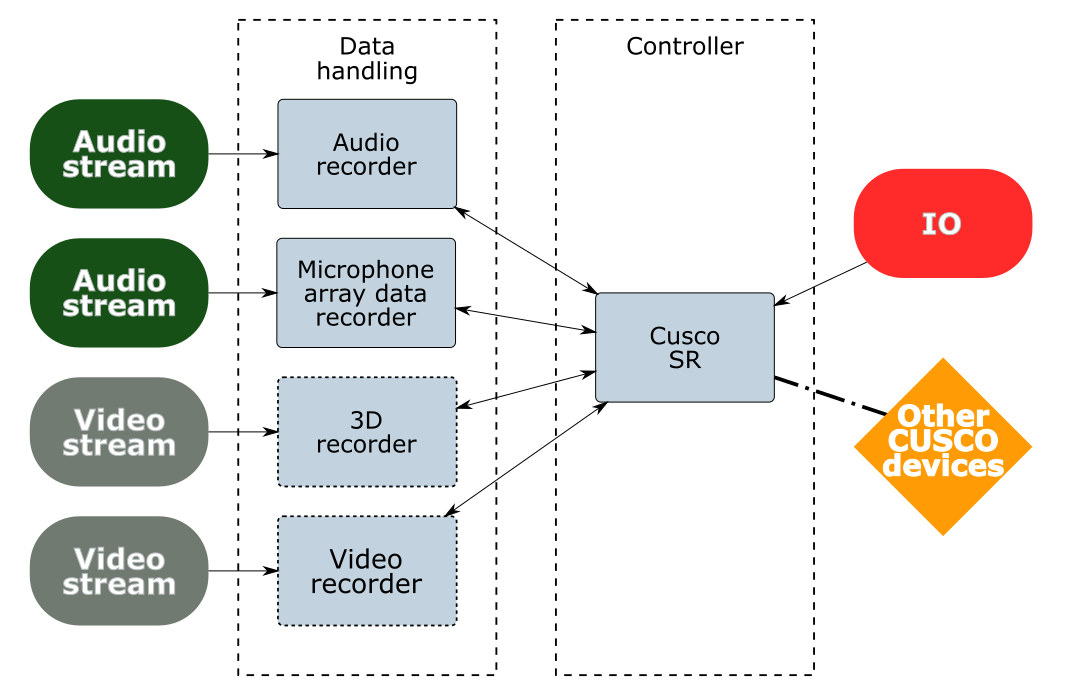}
	\caption{Main components of the CUSCO device.}
	\label{fig:CUSCOarchi}
\end{figure}

Furthermore, data collections can have stricter ethical controls: researchers may also be to prohibited direct access to sensitive information. The device provides capabilities for the collection of anonymised audio and visual features, i.e. an abstracted indirect description of the interaction that cannot be used to reconstruct the original signal.

The software of the device itself is organised around a modular design, as described in Figure \ref{fig:CUSCOarchi}. Each stream, corresponding to a modality (video, audio, 3D) or a function (Voice Activity Detection) is controlled by a dedicated module in charge of setting the configuration, checking the state of required elements (presence of the appropriate device), and managing the recording.

\begin{figure}[htb]
	\centering    
	\includegraphics[width=1\linewidth]{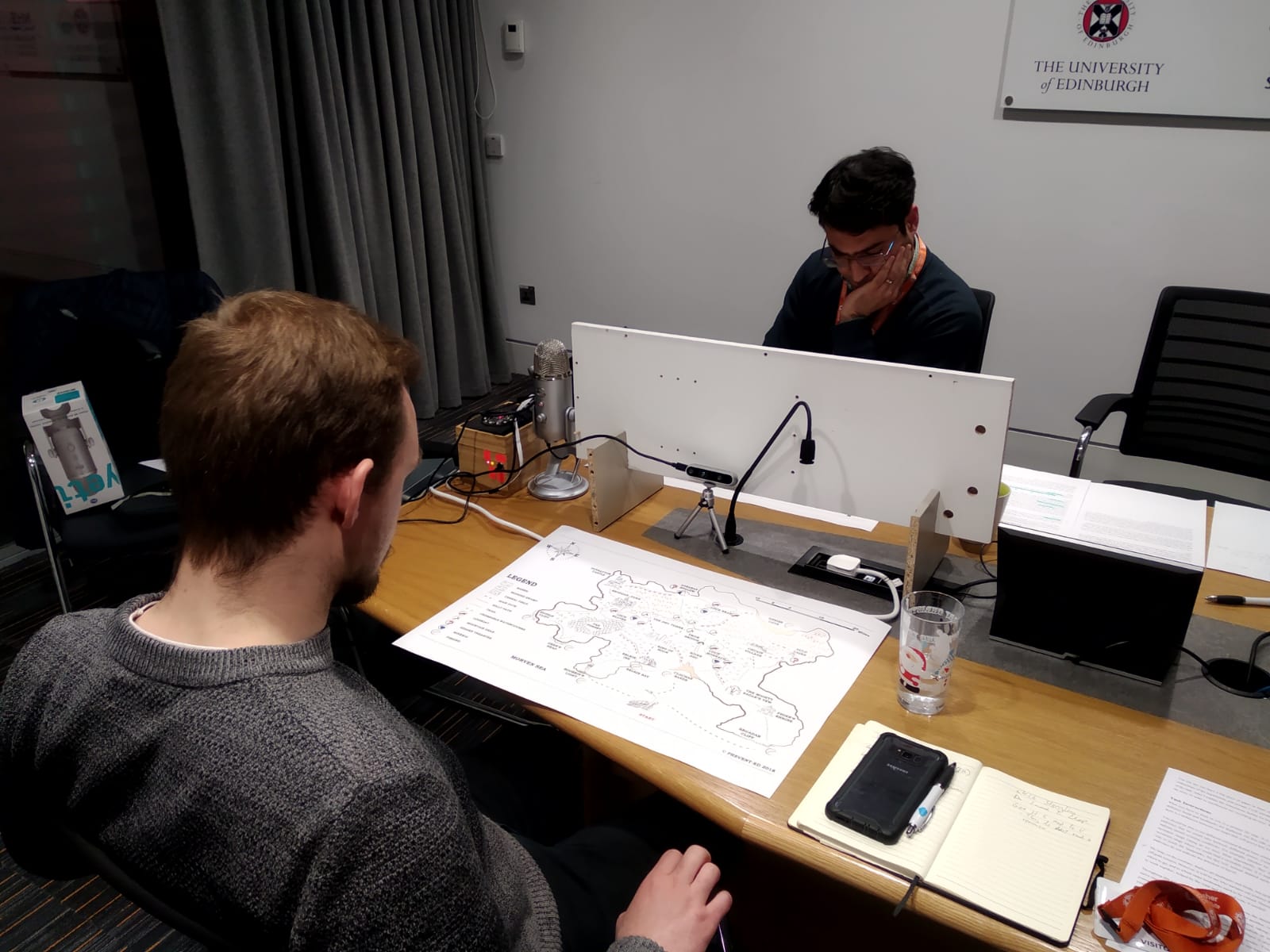}
	\caption{Setup of the recording system.}
	\label{fig:RecordingSetup}
\end{figure}

For the use-case described in this paper (depicted in Figure \ref{fig:RecordingSetup}), we are using two depth cameras, a high-quality table microphone, and a microphone array to facilitate speaker diarisation in post-processing (segmentation of the audio and attribution to each speaker). The decision to use a table microphone was taken because lapel microphones need to be attached to the participants, which can be invasive and can cause distress.
The necessity to record high-quality data (use of an additional microphone and recording close-up video of both participants) lead us to reach the capacity of the system and therefore set multiple devices in a network, two in our case.

The hardware of the device uses common off-the-shelf elements, while the software is open-source. Design schemes and software have been made available online\footnote{\url{https://cybermat.tardis.ed.ac.uk/pial/inca}}.
The need for such a device extends beyond the conversations that we detail in this paper to any sensitive recordings that should be encrypted live. Such use cases include recordings of: GP consultations, interactions with children, and discussions with private companies.

\subsection{Data Handling and Sharing} \label{sub:data}

Once the conversations have been recorded securely, they remain encrypted on the system detailed in section \ref{sub:device} The research team then need to remove any personal information that may have been disclosed during the conversations. To do this, the audio is silenced and the video blurred around the mouth whenever sensitive information is uttered. Blurring video reduces the accuracy of visual behaviour annotation \cite{lasecki2015exploring} but privacy takes precedence to avoid possible participant identification. The transcription can therefore not contain any sensitive information (and should not be transcribed from the original recordings to ensure this).

Personal information will not be shared and it is important to highlight this in the PIS discussed in section \ref{sub:consent} These processed recordings are now considered anonymous as the participants are only identifiable by personal contacts (thus, an unknown researcher cannot identify the participant). The contact details of a member in the research team should be included in the PIS to allow the request for deletion, and subsequent removal, of a participant's data.

The anonymised recordings and associated transcriptions can be shared with other relevant researchers through centralised archives to control its use \cite{derry2010conducting}, \textit{if} stated in the PIS, and results published in research papers. We have decided to store our corpus in DementiaBank \cite{becker1994natural} as it is a shared database of multimedia interactions for the study of communication in dementia. Access to the data in DementiaBank is password protected and restricted to members of the DementiaBank consortium group. Researchers that would benefit from access to this data can request to join this group and therefore benefit from the corpus.

\section{Conclusion}

Collecting multi-modal spontaneous conversations from people with cognitive impairments is a vital step towards creating more accessible and natural SDSs. To ensure this is done ethically, there are many factors that need to be considered which we have collated and detailed throughout Section \ref{sec:ethics} This practical ethical framework can assist researchers who want to navigate the many ethical challenges in order to collect and release corpora of multi-modal interactions. Additionally, CUSCO can be used to securely capture, transport and exchange this data.

\section{Bibliographical References}\label{reference}
%\label{main:ref}

\bibliographystyle{lrec}
\bibliography{lrec2020}

% \section{Language Resource References}
% \label{lr:ref}
% \bibliographystylelanguageresource{lrec}
% \bibliographylanguageresource{languageresource}

\end{document}